\documentclass[
]{ceurart}

\usepackage{hyphenat}
\usepackage{listings}
\usepackage{booktabs}
\usepackage{multirow}
\usepackage{pifont}
\usepackage{mathrsfs}
\newcommand{\cmark}{\ding{51}} 
\newcommand{\xmark}{\ding{55}} 

\definecolor{OurColor}{HTML}{DCEAF7}
\begin{document}

\copyrightyear{2022}
\copyrightclause{Copyright for this paper by its authors.
  Use permitted under Creative Commons License Attribution 4.0
  International (CC BY 4.0).}

\conference{IRCDL'26: 22nd Conference on Information and Research Science Connecting to Digital and Library Science, February 19--20, 2026, Modena, IT}

\title{Harnessing Self-Supervised Features for Art Classification}

\author[1]{Federico Melis}[email=287301@studenti.unimore.it]
\fnmark[1]

\author[1]{Davide Bilardello}[email=285039@studenti.unimore.it]
\fnmark[1]

\author[1]{Emanuele Prato}[email=284215@studenti.unimore.it]
\fnmark[1]

\author[1]{Evelyn Turri}[email=evelyn.turri@unimore.it]

\author[1]{Lorenzo Baraldi}[email=lorenzo.baraldi@unimore.it]

\address[1]{University of Modena and Reggio Emilia, Modena, Italy}
\fntext[1]{These authors contributed equally.}

\begin{abstract}
Classifying artworks presents a significant challenge due to the complex interplay of fine-grained details and abstract features that condition the style or genre of an artwork. This paper presents a systematic investigation of the effectiveness of supervised and self-supervised backbones as feature extractors for both artwork classification and retrieval, with a particular focus on paintings. We conduct an extensive experimental evaluation using the DINO family and CLIP models, assessing multiple classification strategies and feature representations. Our results demonstrate that employing a self-supervised backbone leads to consistent improvements in artwork classification performance. Moreover, our work provides insights into the applicability of classification and retrieval modules in real-world applications, such as virtual reality (VR) applications that support museum navigation.

\end{abstract}

\begin{NoHyper}
\maketitle
\end{NoHyper}

\section{Introduction}
Museums are increasingly adopting digital technologies to enrich visitor engagement with cultural heritage. In particular, Augmented Reality (AR) and Virtual Reality (VR) applications have emerged as significant enhancers for the overall experience in museums. These applications rely on structured digital information to support interactive narratives and personalized navigation. In museums centered on static artworks, such as paintings, having an immersive interaction becomes challenging, and AR applications often depend primarily on their metadata. Accurate style and genre classification is a crucial task in the field, facilitating thematic exploration and recommendations of related artworks based on visual similarity.

Recent advances in deep learning methodologies have demonstrated remarkable performance across various domains. In particular, Convolutional Neural Network (CNN) architectures \cite{resnet, vgg, googlenet} have been highly successful, advancing research in tasks such as object detection and classification, making them more efficient and reliable. However, tasks such as artistic style or genre recognition pose additional challenges: labels are subjective, class boundaries are subtle, and datasets are long-tailed. These factors limit the effectiveness of a fully supervised approach, making it prone to overfitting and reducing its generalization capability. Self-supervised learning (SSL) is particularly effective in disentangling the training objective from the downstream task, capturing more transferable visual structures, and providing robust feature representations. To address the challenge of style and genre classification for artworks, where the required classification is highly specific, we leverage DINO \cite{Caron_2021_ICCV, oquab2023dinov2, simoni2025dinov3} and CLIP \cite{clip} as feature extractors, enabling the capture of general and subtle visual patterns that are less dependent on task-specific labels.

This study presents a comprehensive analysis of self-supervised visual representations for artistic style recognition. Specifically, we examine pre-trained vision encoders whose training objectives are decoupled from the downstream task, such as DINO \cite{Caron_2021_ICCV, oquab2023dinov2, simoni2025dinov3} and CLIP \cite{clip}. We analyze several models on different strategies, \textit{i.e.} zero-shot inference, KNN zero-shot, and linear classification, highlighting the importance of decoupling feature extraction from the classification stage, and providing a clearer understanding of how this design choice affects stylistic classification performance.

Furthermore, these components can be easily integrated into real-world applications. For instance, both classification and retrieval components can support AR-based museum navigation systems and curator tools, assisting museum professionals in categorizing previously unlabeled paintings by style and genre.

We summarize the main contributions of this work as follows:
\begin{itemize}
  \item We demonstrate the effectiveness of self-supervised feature extractors for artistic classification on the WikiArt dataset \cite{wikiart}, highlighting the critical role of decoupled features.
  \item We provide a systematic evaluation of three different classification strategies for self-supervised models, including a comparison with EfficientNetV2 \cite{tan2021efficientnetv2smallermodelsfaster}, used as a supervised baseline.
  \item We perform a retrieval study that highlights the quality of the vision features of our best self-supervised model for the task of artistic style and genre recognition.
\end{itemize}

\section{Related Work}

\paragraph{WikiArt and Artistic Classification.}
The WikiArt \cite{wikiart} dataset offers a diverse benchmark for investigating the classification of artistic attributes, including style, genre, and artist. Early studies \cite{karayev2013recognizing, arora2012towards, zujovic2009classifying} highlight that the wide range of artistic expression poses a significant challenge for traditional computer vision methods. More recent approaches leverage deep learning \cite{cetinic2018fine, artpaintingdetection}, achieving improved performance in stylistic classification and offering interesting insights into how visual representations can bridge the gap between human-level understanding of art while capturing subtle artistic details.
Despite these advances, classifying artworks by style and genre remains a persistent challenge, largely due to the inherently abstract and subjective nature of these concepts. 
Most existing works focus on fine-tuning models pre-trained on large-scale datasets and adapting the extracted features to the artistic domain. Cetinic \textit{et al.} \cite{cetinic2018fine} propose a fine-tuning stage on a convolutional neural network for fine art classification. In contrast, the application of self-supervised and weakly-supervised learning for art classification leaves room for further research. CLIP-Art \cite{Conde_2021_CVPR} proposes to use CLIP \cite{clip} vision features in a zero-shot manner for artistic classification, leveraging natural language information after fine-tuning on the iMet dataset \cite{imet48256}. However, it remains unclear how self-supervised features transfer knowledge to stylistic classification. In this work, we address this gap by investigating performance differences across a wide range of models. 

\paragraph{Self-Supervised Learning.}
The increasing dimension of models and the discoveries in the field of scaling laws \cite{scalinglaws} have amplified the demand for large-scale datasets. However, in the vision domain, annotating such datasets is often prohibitively expensive or even infeasible. Consequently, significant research has focused on developing methods that achieve strong performance without relying on human-labeled data.
Self-supervised learning aims to train a model without explicit training on labeled targets. Weakly-supervised approach, such as CLIP \cite{clip}, leveraging contrastive learning to operate in a high-dimensional space by minimizing the distance between positive pairs while maximizing it for negative pairs. 
On the other hand, self-supervised methods such as BYOL \cite{byol} employ a siamese architecture in which an online network predicts the representation produced by a target network for a different augmented view of the same image. Similarly, DINO \cite{Caron_2021_ICCV} adopts a student–teacher self-distillation paradigm to align representations for different augmentations. Features extracted from DINO models have proven to be effective across multiple downstream tasks. Pérez-García \textit{et al.} \cite{Perez-Garcia2025} show comparable or better performance than biomedical-language supervised models by attaching task-specific decoder heads to a pre-trained DINOv2 \cite{oquab2023dinov2} backbone. Several works also exploit only the visual representations of CLIP, omitting the textual component. For example, CLIP Fusion \cite{ccokmez2025clip} leverages multi-scale semantic features from the pre-trained CLIP visual backbone, in order to define a novel metric for detecting video frame interpolation artifacts. Wu \textit{et al.} \cite{wu2023feature} adapt the CLIP vision encoder for few-shot classification by training an adapter on the original CLIP visual–text embeddings with an additional MLP classifier, combining similarity and classification losses to enhance class separation with limited data.

\section{Approach}

\begin{figure}[t]
    \centering
    \includegraphics[width=0.96\linewidth]{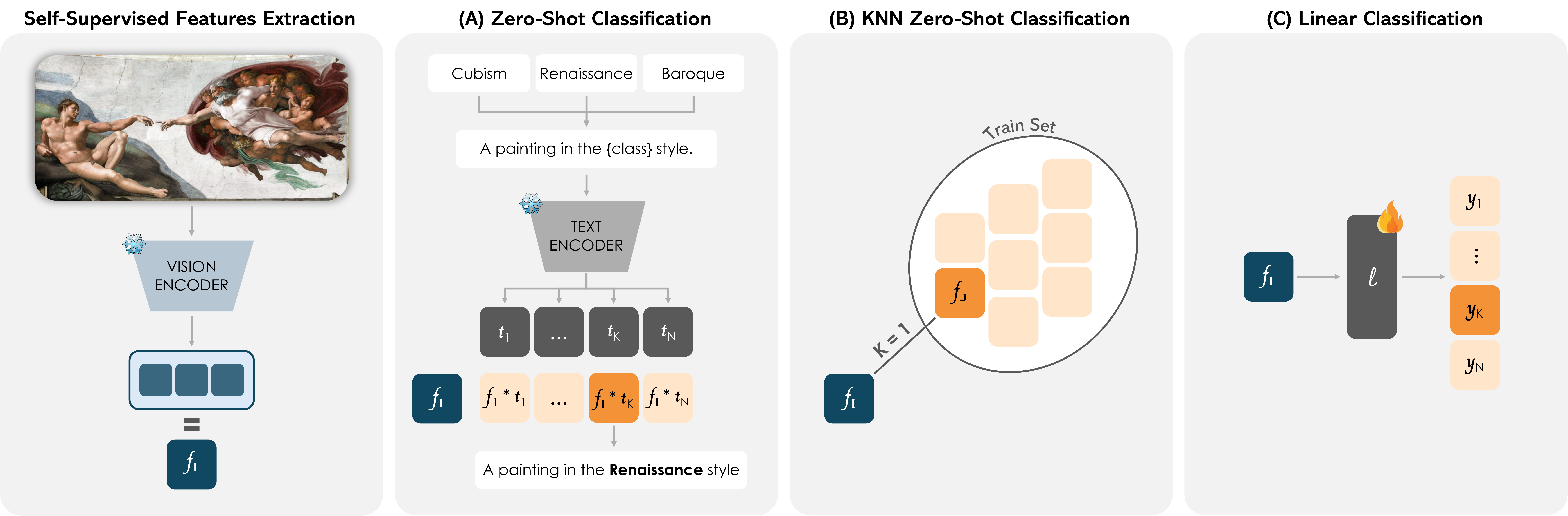}
    \caption{ The figure shows the self-supervised feature extraction phase and each of the proposed classification methods: (A) Zero-Shot Classification, (B) KNN Zero-Shot Classification and (C) Linear Classification.}
    \label{fig:method}
\end{figure}

\subsection{Task Description}
We address two related artistic recognition tasks: artistic style classification and artistic genre classification. Style classification assigns each painting to a specific art style, reflecting characteristic visual conventions, compositional patterns, and historical influences, whereas genre classification emphasizes the thematic content of the artwork, such as landscape, portrait, or illustration. These tasks differ fundamentally from classical image classification, where class labels correspond to well-defined objects or concepts explicitly visible in the image. In contrast, artistic style and genre require the extraction of more abstract features that capture underlying artistic movements or thematic categories. Our experiments provide a comprehensive overview of how SSL models with different representational capabilities adapt to these unconventional and highly nuanced classification challenges.

\subsection{Features Extractors}
We divide the feature extractors analyzed in the work into two groups, \textit{i.e.} supervised and self-supervised. In both cases, we denote the input image as $I$, the visual backbone as $\mathcal{V}$, and, where applicable, the text encoder as $\mathcal{T}$. The visual encoder produces features denoted by $f$, while the text encoder generates features denoted by $t$. Finally, the label set is denoted as $\mathcal{Y} = (y_1, \dots ,y_N)$ where $N$ indicates the number of possible classes. 

\paragraph{Supervised Baseline.}
We employ an EfficientNetV2 \cite{tan2021efficientnetv2smallermodelsfaster} as a reference for the supervised-based model. Unlike the self-supervised models used for the downstream tasks, this baseline is trained end-to-end specifically for the target task. A linear classification layer is attached to the final pooled embedding and trained separately for the specific tasks of style and genre classification. The visual backbone is pre-trained on ImageNet \cite{imagenet}, and jointly fine-tuned during the training of the classification head. This setup reflects a standard supervised transfer-learning approach, allowing the network to adapt its internal knowledge to the artistic domain. The resulting model serves as a benchmark for evaluating the performance of self-supervised features under different downstream strategies.

\paragraph{Self-Supervised Features Extractors.}
For our self-supervised pipeline, we employ a pre-trained vision encoder from CLIP \cite{clip}, and DINO \cite{Caron_2021_ICCV, oquab2023dinov2, simoni2025dinov3}. These models are known to have semantically rich and transferable embeddings, thanks to their training strategy on large datasets. 
To perform our tasks, we use the classification token (CLS) as the extracted feature representation, following common practice in the literature on vision transformer models for classification tasks. We denote as $f_I$ the representation features in output by $\mathcal{V}(I)$. 
Using a frozen encoder and the CLS single global representation ensures generalization for the two classification tasks, enabling comparison across classification strategies.

\subsection{Classification Strategies}
\label{sec:method_strategies}
We implement three distinct approaches for classification using representations extracted from the self-supervised feature extractors. In detail, two training-free approaches, \textit{i.e.} zero-shot and zero-shot KNN classification, and a linear classification technique. The zero-shot KNN and linear classification methods rely solely on the visual backbone $\mathcal{V}$, whereas the zero-shot method leverages both the visual encoder $\mathcal{V}$ and the text encoder $\mathcal{T}$. Figure \ref{fig:method} illustrates the feature extractor pipeline along with the employed classification strategies: (A) Zero-Shot Classification, (B) KNN Zero-Shot Classification and (C) Linear Classification. 

\paragraph{Zero-Shot Classification.}
The first approach is zero-shot classification, which evaluates the model's ability to assign labels to images without task-specific training. In particular, given an image $I$, a label set $\mathcal{Y}=(y_1, ..., y_N)$ with $N$ the number of classes, a visual encoder $ \mathcal{V}$, and a text encoder $\mathcal{T}$, zero-shot classification proceeds by first extracting visual and text features as:
\begin{equation}
    f_I = \mathcal{V}(I) \ \ \ \text{and} \ \ \ t_{k} = \mathcal{T}(p_k) \ \forall \ k \in \{1, ..., N\}
    \label{eq:visual_features}
\end{equation}
where $p_k$ is the textual prompt corresponding to label $y_k$, and $t_{k}$ is the resulting text feature vector. The predicted class $\hat{y}$ is then determined as the label whose textual embedding has the highest cosine similarity with the visual feature $f_I$.
\paragraph{Zero-Shot KNN Classification.}
The second classification approach is zero-shot k-Nearest Neighbors (KNN), which similarly requires no training or fine-tuning. 
This method assigns to a query image $I \in \mathcal{D}_{Query}$ the class of the most similar image within a reference set $\mathcal{D}_{Ref}$. Given the reference set $\mathcal{D}_{Ref}$ and the query set $\mathcal{D}_{Query}$, we extract visual features for the query image $I$ and for each reference image $j \in \mathcal{D}_{Ref}$ as
\begin{equation}
    f_I = \mathcal{V}(I) \ \ \ \text{and} \ \ \ f_j = \mathcal{V}(j) \ \ \forall \ j \in \mathcal{D}_{Ref}.
\end{equation}
The predicted class $\hat{y}$ is then assigned as the label of the reference embedding whose feature vector exhibits the highest cosine similarity with the query embedding $f_I$.

\paragraph{Linear Classification.}
The third classification approach introduces a trainable linear layer $\ell$ applied to the frozen pre-trained visual backbone $\mathcal{V}$, enabling the model to learn a classification mapping from the generalized image representation produced by $\mathcal{V}$. Given the feature vector $f_I$ defined in Equation \ref{eq:visual_features}, the linear classifier outputs a vector $z \in \mathbb{R}^N$, where the predicted class $\hat{y}$ corresponds to the index of the maximum value in $z$.

\subsection{Retrieval}
To obtain a qualitative assessment of the self-supervised feature representations, we implement a retrieval module based on cosine similarity. Visual embeddings for all images in the dataset are extracted using the visual encoder $\mathcal{V}$ as defined in Equation \ref{eq:visual_features}, and these embeddings are indexed using FAISS \cite{douze2024faiss}. Given a query image $I$ with its corresponding embedding $f_I$, the system retrieves the top-$K$ most similar images from the index.

\section{Experiments}
\subsection{Dataset}
We utilize the WikiArt dataset \cite{wikiart}, a collection of approximately 80,000 paintings, encompassing 27 style classes and 11 genre classes. Examples of styles include Impressionism, Cubism, Baroque, and Romanticism, while genres cover categories such as illustration, abstract painting, landscape, and portrait. The genre taxonomy also includes the ``Unknown Genre'' label, which we remove from all the experiments. The dataset is partitioned into train (80\%), validation (10\%), and test (10\%) splits. The validation set is used exclusively for baseline and for the linear classifier training, while train and test sets are fixed across all the experiments. 

\subsection{Implementation Details}
We detail the experimental setup for the three classification strategies in the following paragraphs to facilitate reproducibility. We report hardware specifications, the hyperparameters used, and a concise summary of methodological details. All experiments were conducted on a single NVIDIA A40 (48 GB). The random seed is fixed to 42 for reproducibility.

\paragraph{Supervised Baseline.}
We employ two separate visual backbones, one for the style classification task and the other for genre classification, both initialized from identical EfficientNetV2-L checkpoints \cite{tan2021efficientnetv2smallermodelsfaster} pre-trained on ImageNet \cite{imagenet}. The default final MLP layer of EfficientNetV2-L is replaced with a linear layer mapping the model’s embedding dimension to the respective number of classes. We apply a dropout \cite{srivastava2014dropout} rate of $0.4$ before the linear layer for regularization. Models are trained for 10 epochs with a batch size of 16, and the best-performing model is selected for evaluation. The full EfficientNetV2-L backbone is fine-tuned using cross-entropy loss with a learning rate of $10^{-5}$ and a weight decay of $10^{-5}$ \cite{loshchilov2018decoupled}, while the classification head is trained with the same loss but a higher learning rate of $10^{-4}$. All optimization is performed using the Adam optimizer \cite{kingma2017adammethodstochasticoptimization}.

\paragraph{Self-Supervised Features Extractors.}
We employ two self-supervised feature extractors: CLIP-ViT-L/14 and DINOV3-ViT-L/16. For the zero-shot classification strategy, CLIP uses its native text encoder, while DINO utilizes the \texttt{dino.txt} \cite{Jose_2025_CVPR}, text encoder specifically aligned with the DINO's visual encoder representation space. Prompts for style and genre classification are constructed as follows: ``\textit{A painting in the <class> style.}'' for style, and ``\textit{A <genre> painting.}'' for genre.
For zero-shot KNN classification, the reference set $\mathcal{D}_{Ref}$ corresponds to the WikiArt training set, and the query set $\mathcal{D}_{Query}$ corresponds to the test set. Linear classification layers are trained using cross-entropy loss, a learning rate of $10^{-4}$, weight decay equal to $10^{-4}$,  and a batch size of 1024. Training proceeds for up to 100 epochs with early stopping (patience = 5).

In all experiments, we refer to models using the zero-shot approach as CLIP Zero-Shot and DINO Zero-Shot, models evaluated with the KNN strategy are denoted as CLIP-KNN and DINO-KNN, and models with a trainable linear layer as CLIP-Linear and DINO-Linear.

\begin{table}[t]
  \centering
  \resizebox{\textwidth}{!}{
    \begin{tabular}{l c cccc cccc}
      \toprule
      \multirow{2}{*}{Model} & \multirow{2}{*}{Trainable}
      & \multicolumn{4}{c}{Style} & \multicolumn{4}{c}{Genre} \\
      \cmidrule(lr){3-6} \cmidrule(lr){7-10}
       & & P & R & F1 & acc@1 & P & R & F1 & acc@1 \\
      \midrule
      EfficientNetV2 & \cmark & 68.9 & \textbf{68.6} & 68.3 & 68.6 & 82.0 & 82.2 & 82.0 & 82.2 \\
      \cmidrule(lr){1-10}
      DINO Zero-Shot & \xmark & 29.1 & 34.8 & 26.1 & 27.4 & 66.9 & 67.5 & 63.7 & 67.2 \\
      CLIP Zero-Shot & \xmark & 39.3 & 41.2 & 36.8 & 41.9 & 69.4 & 65.6 & 60.2 & 64.4 \\
      DINO-KNN & \xmark & 63.5 & 62.4 & 61.7 & 63.2 & 78.5 & 78.0 & 78.1 & 80.2 \\
      CLIP-KNN & \xmark &  69.2 & \textbf{68.6} & 68.3 & \textbf{70.6} & 80.3 & 80.9 & 80.5 & 81.7 \\
      DINO-Linear & \cmark & 71.5 & 61.4 & 65.0 & 65.0 & 81.6 & 80.6 & 81.0 & 83.6 \\
      \rowcolor{OurColor}
      CLIP-Linear (\textbf{Ours}) & \cmark & \textbf{74.3} & 67.2 & \textbf{69.8} & 69.2 & \textbf{83.2} & \textbf{82.5} & \textbf{82.8} & \textbf{84.9} \\
      \bottomrule
    \end{tabular}
  }
  \caption{Comparison of the baseline EfficientNetV2 \cite{tan2021efficientnetv2smallermodelsfaster}, compared to CLIP-ViT-L/14 \cite{clip} and DINOV3-ViT-L/16 \cite{simoni2025dinov3} with each classification strategy: Zero-Shot, KNN Zero-Shot, and Linear classification. The table reports precision (P), recall (R), F1-score (F1), and top-1 accuracy (acc@1) for both style and genre prediction, and highlights whether additional trainable components are used. All the results are computed on the test set.} 
  \label{tab:comparative-results}
\end{table}
\subsection{Performance on WikiArt}
\paragraph{Classification.}
\begin{figure}[t]
    \centering
    \includegraphics[width=0.96\linewidth]{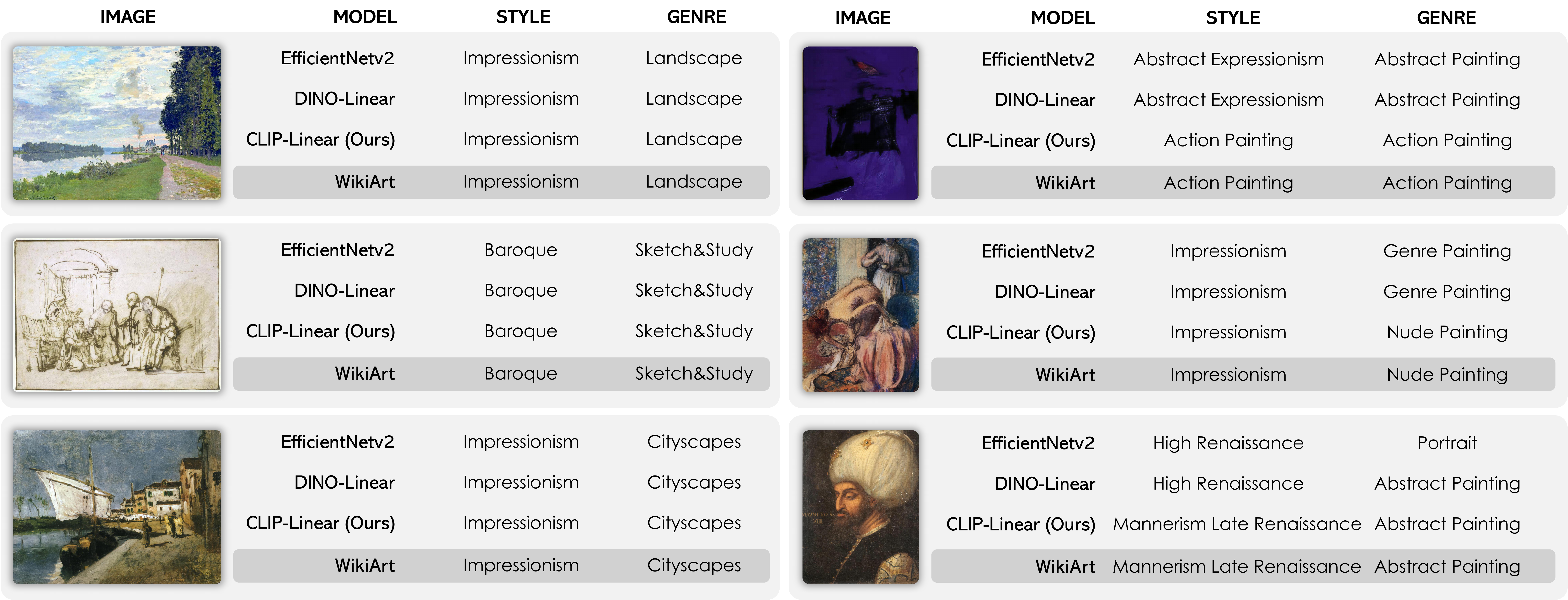}
    \caption{Qualitative results for style and genre classification. For each image, we present predictions for EfficientNetV2, DINO-Linear and CLIP-Linear (Ours) compared to WikiArt \cite{wikiart} Ground Truth.}
    \label{fig:qualitative}
\end{figure}

We report in Table~\ref{tab:comparative-results} the results achieved under the best configuration for each method, explicitly indicating whether additional trainable components are introduced. The finetuned EfficientNetV2-L with a newly trained classification head achieves solid overall performance on both style and genre classification. Notably, DINO-KNN reaches comparable performance to the finetuned EfficientNetV2-L, despite requiring neither additional trainable parameters nor any finetuning phase. Linear classification settings, instead, outperform all the other configurations, benefiting from high-quality visual features combined with a supervised classification head. Although DINO features are generally recognized for capturing rich semantic information, our experiments reveal consistently stronger performance from CLIP across all evaluation protocols. This suggests that style and genre classification rely more heavily on nuanced visual features than on broad semantic structure. In particular, CLIP-KNN outperforms the baseline across all style metrics, whereas the corresponding DINO-KNN model does not. Best performance is then achieved with CLIP-Linear, indicating that features extracted from the frozen backbone are rich and highly discriminative for both tasks, and that a simple linear layer can easily map those features into each class. It is also noteworthy that zero-shot setting, both for DINO and CLIP models, performs poorly, implying that textual prompts provide limited benefit for style and genre recognition, and that these tasks are primarily driven by visual rather than textual alignment.

The qualitative results in Figure~\ref{fig:qualitative} illustrate six examples comparing Ground Truth, the baseline, DINO-Linear, and CLIP-Linear. The experiments include the strongest configuration for each self-supervised feature extractor. In the three examples on the left, both baseline, DINO-Linear, and CLIP-Linear correctly predict both style and genre classes. In the images on the right, instead, only CLIP-Linear correctly predicts style and genre, while both the baseline and DINO-Linear exhibit failures on either style or genre classification.

\begin{figure}[t]
    \centering
    \includegraphics[width=0.94\linewidth]{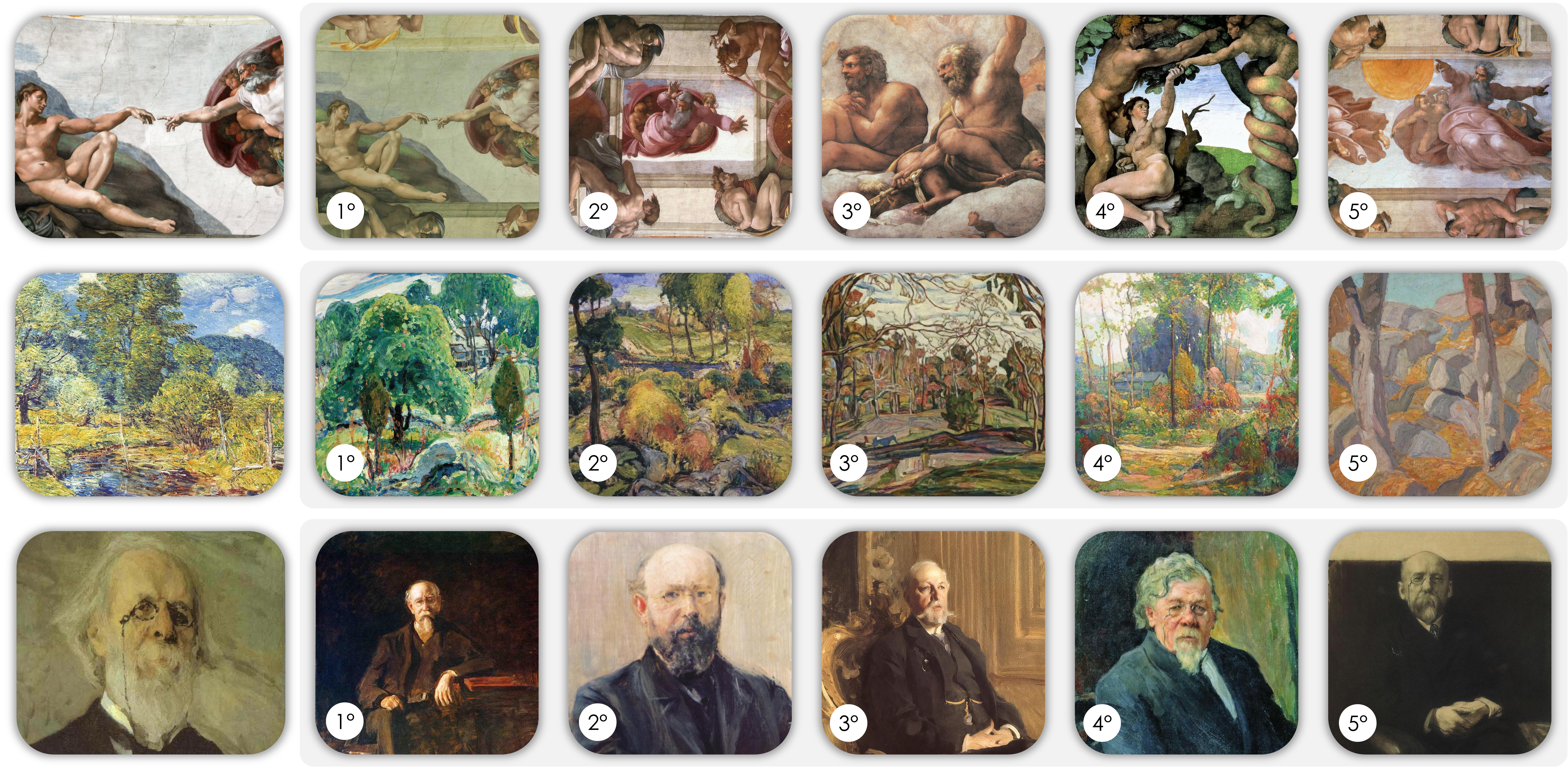}
    \caption{This figure shows qualitative retrieval results obtained with CLIP-ViT-L/14. The query images are depicted in the first column for each row, and they span across different classes of style and genre. For each query sample, we show top-5 best retrieved images.}
    \label{fig:retrieval}
\end{figure}

\paragraph{Retrieval.}
To further assess the semantic richness and discriminative power of the representations extracted by our strongest self-supervised model, CLIP, we conduct a series of qualitative retrieval experiments on the WikiArt \cite{wikiart} dataset. Figure \ref{fig:retrieval} shows three different qualitative examples for retrieval. In details for each row, the image on the left serves as the query for the retrieval stage, while the numbered pictures are the ones retrieved using the strategy explained in Section \ref{sec:method_strategies}. In all three examples, the retrieved images are coherent with the query. In particular, in all three cases, both style and genre are similar to the main picture. In addition, it is clear that the subject is also coherent with the reference. The final row offers a particularly clear illustration: the retrieved paintings not only share the same stylistic and genre characteristics but also exhibit strongly similar compositional features, such as the pose, hair, beard, and glasses of the subject. To highlight the robustness and generality of these features, we include an additional retrieval example using a query image outside the WikiArt dataset. The query depicts "\textit{La creazione di Adamo}" by Michelangelo. Remarkably, the top retrieved image from WikiArt corresponds to the same artwork, despite clear differences in color palette and rendering. CLIP successfully identifies the match by relying on deeper visual and semantic cues rather than superficial appearance. The remaining retrieved images also maintain strong consistency with the query in both style and genre, mirroring the behavior observed in the other examples.

\section{Conclusion}
In this work, we present a systematic overview of self-supervised learning backbones applied to the challenging task of classifying artistic style and genre. By benchmarking diverse approaches, we demonstrate that leveraging pre-trained features from DINO and CLIP significantly benefits artwork classification. Our experiments highlight the value of decoupling feature extraction from the downstream classification task. Most surprisingly, this held true for the CLIP-Linear setup, which performed better than full fine-tuning of the EfficientNetV2 baseline for both style and genre classification. This suggests that multi-modal pre-training in CLIP captures more abstract artistic aspects compared to standard visual supervision. Moreover, the zero-shot KNN strategy proves to be remarkably competitive, achieving performance comparable to supervised methods without any additional training. This further validates the intrinsic semantic quality of self-supervised visual features verified by our qualitative retrieval analysis.

These findings have direct implications for the cultural heritage sector. The explored architectures and strategies are particularly well-suited for integration into immersive AR and VR museum experiences, enabling context-aware navigation and dynamic artwork recommendation for visitors. Additionally, the use of pre-trained features opens opportunities for scalable applications in large-scale digital archives, where manual labeling is often infeasible, supporting more adaptive organization of art collections. The approach also has the potential to enhance accessibility, for instance by providing visually enriched navigation systems for audiences with diverse needs. Finally, it offers educational value, allowing students and learners to explore art collections by style or genre, thereby improving engagement and understanding of artistic heritage.

\bibliography{sample-ceur}

@string{cvpr     = {CVPR}}

@string{cvprw    = {CVPR Workshops}}

@string{iccv     = {ICCV}}

@string{iclr     = {ICLR}}

@string{icml     = {ICML}}

@string{icpr     = {ICPR}}

@string{wacv     = {WACV}}

@string{ieeetip  = {IEEE TIP}}

@inproceedings{Jose_2025_CVPR,
  title = {{DINOv2 Meets Text: A Unified Framework for Image- and Pixel-Level Vision-Language Alignment}},
  author = {Jose, Cijo and Moutakanni, Th\'eo and Kang, Dahyun and Baldassarre, Federico and Darcet, Timoth\'ee and Xu, Hu and Li, Daniel and Szafraniec, Marc and Ramamonjisoa, Micha\"el and Oquab, Maxime and Sim\'eoni, Oriane and Vo, Huy V. and Labatut, Patrick and Bojanowski, Piotr},
  booktitle = cvpr,
  year = {2025}
}

@article{byol,
  title = {{Bootstrap Your Own Latent - A New Approach to Self-Supervised Learning}},
  author={Grill, Jean-Bastien and Strub, Florian and Altch{\'e}, Florent and Tallec, Corentin and Richemond, Pierre and Buchatskaya, Elena and Doersch, Carl and Avila Pires, Bernardo and Guo, Zhaohan and Gheshlaghi Azar, Mohammad and others},
  journal={Advances in neural information processing systems},
  year={2020}
}

@article{scalinglaws,
  title = {{Scaling Laws for Neural Language Models}},
  author = {Kaplan, Jared and McCandlish, Sam and Henighan, Tom and Brown, Tom B and Chess, Benjamin and Child, Rewon and Gray, Scott and Radford, Alec and Wu, Jeffrey and Amodei, Dario},
  journal = {arXiv preprint arXiv:2001.08361},
  year = {2020}
}

@inproceedings{clip,
  title={{Learning Transferable Visual Models From Natural Language Supervision}},
  author={Radford, Alec and Kim, Jong Wook and Hallacy, Chris and Ramesh, Aditya and Goh, Gabriel and Agarwal, Sandhini and Sastry, Girish and Askell, Amanda and Mishkin, Pamela and Clark, Jack and others},
  booktitle=icml,
  year={2021},
}

@article{simoni2025dinov3,
  title={{Dinov3}},
  author={Sim{\'e}oni, Oriane and Vo, Huy V and Seitzer, Maximilian and Baldassarre, Federico and Oquab, Maxime and Jose, Cijo and Khalidov, Vasil and Szafraniec, Marc and Yi, Seungeun and Ramamonjisoa, Micha{\"e}l and others},
  journal={arXiv preprint arXiv:2508.10104},
  year={2025}
}

@article{oquab2023dinov2,
  title={{{DINO}v2: Learning Robust Visual Features without Supervision}},
  author={Maxime Oquab and Timoth{\'e}e Darcet and Th{\'e}o Moutakanni and Huy V. Vo and Marc Szafraniec and Vasil Khalidov and Pierre Fernandez and Daniel Haziza and Francisco Massa and Alaaeldin El-Nouby and Mido Assran and Nicolas Ballas and Wojciech Galuba and Russell Howes and Po-Yao Huang and Shang-Wen Li and Ishan Misra and Michael Rabbat and Vasu Sharma and Gabriel Synnaeve and Hu Xu and Herve Jegou and Julien Mairal and Patrick Labatut   and Armand Joulin and Piotr Bojanowski},
  journal={Transactions on Machine Learning Research},
  year={2024}
}

@inproceedings{loshchilov2018decoupled,
  title={{Decoupled Weight Decay Regularization}},
  author={Ilya Loshchilov and Frank Hutter},
  booktitle=iclr,
  year={2019}
}

@inproceedings{googlenet,
  title = {{Going Deeper With Convolutions}},
  author={Szegedy, Christian and Liu, Wei and Jia, Yangqing and Sermanet, Pierre and Reed, Scott and Anguelov, Dragomir and Erhan, Dumitru and Vanhoucke, Vincent and Rabinovich, Andrew},
  booktitle=cvpr,
  year={2015}
}

@inproceedings{resnet,
  title = {{Deep Residual Learning for Image Recognition}},
  author={He, Kaiming and Zhang, Xiangyu and Ren, Shaoqing and Sun, Jian},
  booktitle=cvpr,
  year={2016}
}

@inproceedings{vgg,
  title={{Very Deep Convolutional Networks for Large-Scale Image Recognition}}, 
  author={Simonyan, K and Zisserman, A},
  booktitle=iclr,
  year={2015},
}

@article{kingma2017adammethodstochasticoptimization,
  title={{Adam: A Method for Stochastic Optimization}},
  author={Diederik P. Kingma and Jimmy Ba},
  journal={CoRR},
  year={2014}
}

@inproceedings{Caron_2021_ICCV,
    title={{Emerging Properties in Self-Supervised Vision Transformers}},
    author={Caron, Mathilde and Touvron, Hugo and Misra, Ishan and J\'egou, Herv\'e and Mairal, Julien and Bojanowski, Piotr and Joulin, Armand},
    booktitle=iccv,
    year={2021},
}

@article{srivastava2014dropout,
  title={{Dropout: A Simple Way to Prevent Neural Networks from Overfitting}},
  author={Srivastava, Nitish and Hinton, Geoffrey and Krizhevsky, Alex and Sutskever, Ilya and Salakhutdinov, Ruslan},
  journal={The journal of machine learning research},
  year={2014},
}

@inproceedings{imagenet,
  title={{Imagenet: A large-scale hierarchical image database}},
  author={Deng, Jia and Dong, Wei and Socher, Richard and Li, Li-Jia and Li, Kai and Fei-Fei, Li},
  booktitle=cvpr,
  year={2009},
}

@inproceedings{tan2021efficientnetv2smallermodelsfaster,
  title = {{EfficientNetV2: Smaller Models and Faster Training}},
  author = {Tan, Mingxing and Le, Quoc},
  booktitle = icml,
  year = {2021}
}

@article{artpaintingdetection,
  title={{Art painting detection and identification based on deep learning and image local features}},
  author={Hong, Yiyu and Kim, Jongweon},
  journal={Multimedia Tools and Applications},
  year={2019},
}

@article{wikiart,
  title={{Improved ArtGAN for Conditional Synthesis of Natural Image and Artwork}}, 
  author={Tan, Wei Ren and Chan, Chee Seng and Aguirre, Hernan E and Tanaka, Kiyoshi},
  journal=ieeetip,
  year={2018},
}

@article{douze2024faiss,
  title={{The Faiss Library}}, 
  author={Douze, Matthijs and Guzhva, Alexandr and Deng, Chengqi and Johnson, Jeff and Szilvasy, Gergely and Mazar{\'e}, Pierre-Emmanuel and Lomeli, Maria and Hosseini, Lucas and J{\'e}gou, Herv{\'e}},
  journal={IEEE Transactions on Big Data},
  year={2025},
  publisher={IEEE}
}

@article{karayev2013recognizing,
  title={{Recognizing Image Style}},
  author={Karayev, Sergey and Trentacoste, Matthew and Han, Helen and Agarwala, Aseem and Darrell, Trevor and Hertzmann, Aaron and Winnemoeller, Holger},
  journal={arXiv preprint arXiv:1311.3715},
  year={2013}
}

@inproceedings{arora2012towards,
  title={{Towards automated classification of fine-art painting style: A comparative study}},
  author={Arora, Ravneet Singh and Elgammal, Ahmed},
  booktitle=icpr,
  year={2012},
}

@inproceedings{zujovic2009classifying,
  title={{Classifying paintings by artistic genre: An analysis of features \& classifiers}},
  author={Zujovic, Jana and Gandy, Lisa and Friedman, Scott and Pardo, Bryan and Pappas, Thrasyvoulos N},
  booktitle={2009 IEEE international workshop on multimedia signal processing},
  year={2009},
}

@article{cetinic2018fine,
  title = {{Fine-tuning Convolutional Neural Networks for fine art classification}},
  author={Cetinic, Eva and Lipic, Tomislav and Grgic, Sonja},
  journal={Expert Systems with Applications},
  year={2018},
}

@inproceedings{Conde_2021_CVPR,
  title = {{CLIP-Art: Contrastive Pre-Training for Fine-Grained Art Classification}},
  author = {Conde, Marcos V. and Turgutlu, Kerem},
  booktitle = cvprw,
  year = {2021},
}

@misc{imet48256,
    title	= {{The iMet Collection 2019 Challenge Dataset}},
    author	= {Chenyang Zhang and Christine Kaeser-Chen and Grace Vesom and Jennie Choi and Maria Kessler and Serge Belongie},
    year	= {2019}
}

@inproceedings{wu2023feature,
  title = {{Feature Adaptation with CLIP for Few-shot Classification}},
  author={Wu, Guangxing and Chen, Junxi and Zhang, Wentao and Wang, Ruixuan},
  booktitle={Proceedings of the 5th ACM International Conference on Multimedia in Asia},
  year={2023}
}

@article{Perez-Garcia2025,
  title={{Exploring scalable medical image encoders beyond text supervision}},
  author={P{\'e}rez-Garc{\'i}a, Fernando and Sharma, Harshita and Bond-Taylor, Sam and Bouzid, Kenza and Salvatelli, Valentina and Ilse, Maximilian and Bannur, Shruthi and Castro, Daniel C. and Schwaighofer, Anton and Lungren, Matthew P. and Wetscherek, Maria Teodora and Codella, Noel and Hyland, Stephanie L. and Alvarez-Valle, Javier and Oktay, Ozan},
  journal={Nature Machine Intelligence},
  year={2025},
}

@inproceedings{ccokmez2025clip,
  title={{CLIP-Fusion: A Spatio-Temporal Quality Metric for Frame Interpolation}},
  author={{\c{C}}{\"o}kmez, G{\"o}ksel Mert and Zhang, Yang and Schroers, Christopher and Aydin, Tun{\c{c}} Ozan},
  booktitle=wacv,
  year={2025},
}

\end{document}